\documentclass[conference]{IEEEtran}
\IEEEoverridecommandlockouts

\usepackage{cite}
\usepackage{amsmath,amssymb,amsfonts}
\usepackage{algorithmic}
\usepackage{graphicx}
\usepackage{textcomp}
\usepackage{xcolor}
\usepackage{multirow}

\def\BibTeX{{\rm B\kern-.05em{\sc i\kern-.025em b}\kern-.08em
    T\kern-.1667em\lower.7ex\hbox{E}\kern-.125emX}}

\newcommand\blfootnote[1]{%
  \begingroup
  \renewcommand\thefootnote{}\footnote{#1}%
  \addtocounter{footnote}{-1}%
  \endgroup
}

\begin{document}

\makeatletter
\newcommand{\linebreakand}{%
  \end{@IEEEauthorhalign}
  \hfill\mbox{}\par
  \mbox{}\hfill\begin{@IEEEauthorhalign}
}
\makeatother

\title{Deep Learning for identifying systolic complexes in SCG traces: a cross-dataset analysis \\

}

\author{\IEEEauthorblockN{Michele Craighero}
\IEEEauthorblockA{\textit{Politecnico di Milano}\\
Milan, Italy \\
michele.craighero@polimi.it}
\and
\IEEEauthorblockN{Sarah Solbiati}
\IEEEauthorblockA{\textit{Politecnico di Milano}\\
Milan, Italy \\
sarah.solbiati@polimi.it}
\and
\IEEEauthorblockN{Federica Mozzini}
\IEEEauthorblockA{\textit{Politecnico di Milano}\\
Milan, Italy \\
federica.mozzini@polimi.it}
\linebreakand
\IEEEauthorblockN{Enrico Caiani}
\IEEEauthorblockA{\textit{Politecnico di Milano}\\
Milan, Italy \\
enrico.caiani@polimi.it}
\and
\IEEEauthorblockN{Giacomo Boracchi}
\IEEEauthorblockA{\textit{Politecnico di Milano}\\
Milan, Italy \\
giacomo.boracchi@polimi.it}
}

\maketitle
\vspace*{-0.4cm}
\begin{abstract}
The seismocardiographic signal is a promising alternative to the traditional ECG in the analysis of the cardiac activity. In particular, the systolic complex is known to be the most informative part of the seismocardiogram, thus requiring further analysis. State-of-art solutions to detect the systolic complex are based on Deep Learning models, which have been proven effective in pioneering studies. However, these solutions have only been tested in a controlled scenario considering only clean signals acquired from users maintained still in supine position. On top of that, all these studies consider data coming from a single dataset, ignoring the benefits and challenges related to a cross-dataset scenario. In this work, a cross-dataset experimental analysis was performed considering also data from a real-world scenario. Our findings prove the effectiveness of a deep learning solution, while showing the importance of a personalization step to contrast the domain shift, namely a change in data distribution between training and testing data. Finally, we demonstrate the benefits of a multi-channels approach, leveraging the information extracted from both accelerometers and gyroscopes data.

\end{abstract}

\vspace{-1mm}
\section{Introduction}
\blfootnote{This paper is partially supported by PNRR-PE-AI FAIR project funded by the NextGeneration EU program.}
The analysis of cardiac activity is key in healthcare, in particular to detect and monitor cardiovascular diseases. Nowadays, the electrocardiographic (ECG) signal represents the gold standard to study the cardiac electrical activity and it is the most used diagnostic tool in hospitals. 
However, ECG measurements are not easily accessible in out-of-clinic environments, since to gather ECG traces in a format typical of medical diagnosis it is necessary to place multiple electrodes in precise sites on the skin. 
On top of that, the ECG signal does not provide direct information on the mechanical function of the heart, which represents the natural and important counterpart to complement the ECG analysis in a diagnosis of a patient with suspected heart disease. 
For these reasons, the analysis of the seismocardoardiogram (SCG) signal has gathered a lot of interest in the last decades. 

The SCG is a mechanical signal that reflects the micromovements of the chest due to the precordial vibrations produced by beating heart \cite{bozhenko, zanettiSCG, landreani}. This signal is typically measured with accelerometer sensors and, therefore, can be easily collected in a out-of-clinic environment. The SCG morphology is characterized by a sequence of deflections, which correspond to mechanical events of the hearth. In particular, the isovolumic contraction (IVC) and the aortic opening (AO) are the most common fiducial points visible in a SCG signal \cite{zanetti}. Starting from these deflections, it is possible to identify the systolic complex within every heartbeat, as shown in Figure \ref{fig:syst}. 

However, the intrinsic mechanical nature of the SCG makes this signal highly susceptible to environmental noise due to user's movements. In particular, the identification of the systolic complex in the SCG, without using the ECG as reference, is not straightforward as expected, as it can be masked by user movements and other types of noise. In addition, the morphology of the SCG signal can significantly change between subjects or being modified due to the respiratory phase and the characteristic deflections of the systolic complex can appear also in the diastolic part. For these reasons, the SCG is typically measured in controlled conditions, having the subject awake very still in supine position.

In the last years, in the literature there have been proposed some ECG-free deep learning (DL) solutions to analyse the SCG signal \cite{suresh, duraj}. In particular, ECG-free solutions adopt the SCG signal only, limiting the role of ECG for generating the ground truth of the training set. The ECG is then excluded, as it is not required during the operational life of the model. Even though reaching promising results, all these works were tested in a simplified scenario, which did not reflect real-world operating conditions. In fact, the experimental analyses in \cite{suresh, duraj} employed a single dataset containing recordings collected in a controlled environment.

To overcome these limitations, in this work we conduct a cross-dataset experimental analysis using DL models to identify systolic complexes in SCG signals without ECG. More in detail, the performance of a state-of-art DL model, namely a U-Net, was analyzed by training/testing models on multiple datasets. In this way, we introduce different levels of domain shift, namely a difference in distributions between training and testing data, reproducing the gap between signals recorded in controlled conditions and in real-world. In fact, DL models pretrained on clean signals from a public dataset could not be effective when analyzing real-world SCG signals, which are recorded in uncontrolled conditions. Our experimental analysis is also meant to assess the benefits of traditional fine-tuning and model personalization. Finally, the contribution of the different channels of accelerometers and gyroscopes was evaluated in our challenging scenario.

In summary, our contributions are as follows:
\begin{itemize}
    \item an extensive experimental analysis on SCG systolic complex identification was conducted, considering for the first time the challenging scenario where data are coming from multiple datasets, thus giving rise to domain shift.
    \item the performance of a state-of-art DL model on real-world data collected in an uncontrolled environment was assessed, analysing also fine-tuning and personalization.
    \item the difference between a multi-channels approach w.r.t. to a single-channel one in our ECG-free systolic complex identification task was evaluated.
\end{itemize}

\begin{figure}
    \centering  \includegraphics[width=0.45\textwidth]{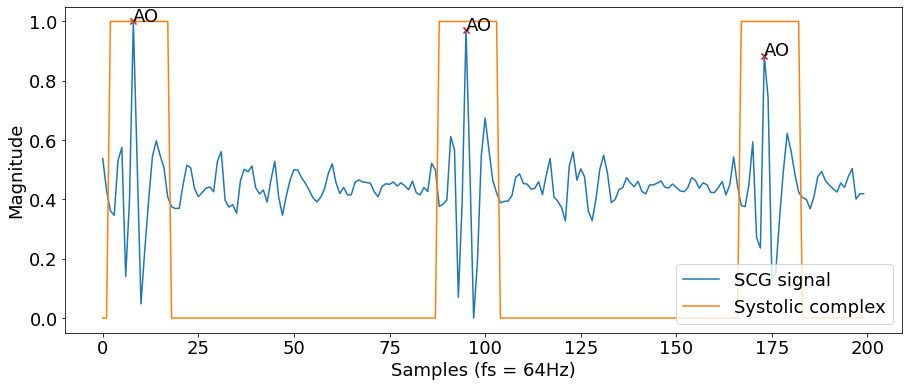}
    \vspace{-3mm}
    \caption{AO fiducial points and systolic complexes in a clean portion of SCG signal. Note that the bounding boxes are centered in the AO points.}
    \label{fig:syst}
    \vspace{-0.54cm}
\end{figure}
\vspace{-1.1mm}
\section{Materials and Methods}
The automatic identification of the systolic complex in the SCG signal is of key importance in the biomedical scenario, as it represents the starting point for any ECG-free cardiac analysis, like the detection of the AO point.
\vspace{-1mm}
\subsection{Datasets}
Our work aims to provide an extensive experimental analysis of the ECG-free systolic complex identification task. For this reason, 3 different datasets were adopted:
\begin{itemize}
    \item CEBS \cite{cebs1,cebs2}: publicly available at \emph{Physionet.org} \cite{PhysioNet}, 20 healthy subjects awake in supine position, 1h recording per subject, fs=5 kHz, only z-axis of accelerometer;
    
    \item MEC, Mechanocardiograms with ECG reference \cite{kaisti, kaistiPaper}: publicly available at \emph{ieee-dataport.org}, 29 healthy subjects awake in supine position, 260min of recordings in total, fs=800Hz, 6 channels (tri-axial accelerometer, tri-axial gyroscope);
    \item BioPoli: private, 23 subjects performing daily-life activities, 24h recording per subject, fs=64Hz, 6 channels (tri-axial accelerometer, tri-axial gyroscope).
\end{itemize}

The 3 datasets differ in the number of subjects, in the sampling frequency and in the number of channels of the SCG signal. 
It is fundamental to underline that the CEBS and MEC datasets contain SCG signals recorded in a controlled environment, while the subject is maintained still in a supine position. On the contrary, the BioPoli dataset contains 24h recordings collected in uncontrolled conditions, thus reflecting a real-world scenario. From the BioPoli dataset we selected only the SCG portions corresponding to the sleeping period, where subjects nevertheless can move and change their position.
 
To generate the ground truth location of the systolic complexes, we pursued the approach described in \cite{Aoidentification} and \cite{duraj}, which leverages a synchronized ECG recording to detect the AO peak of the SCG within every heartbeat. We clarify that the ECG signal is used only in this labeling step, while the training and evaluation are completely ECG-free. As first step, the Pan-Tompkins algorithm localizes all the R-waves in the ECG signal. Then, the R-wave was utilized as the starting point for the identification of the aortic valve opening on the SCG, which indicates the start of the ventricular contraction. In particular, the AO fiducial point on the SCG was located as the local maximum within a 90 ms window after the R-wave of the ECG signal.

To localize the systolic complexes, we defined a 25ms bounding box around each AO fiducial point, which served as ground-truths for our network. As shown in Figure \ref{fig:syst}, bounding boxes contain all the characteristic deflections of the SCG systolic complex. 

\subsection{SCG Systolic Identification algorithm}
As first step, training data were preprocessed by splitting every SCG signal in windows of 5 seconds, so that each window contains more than one heartbeat. Each window was normalized to a [0, 1] interval using a min-max normalization. These steps preserve the signal morphology and limit the impact of outliers to the corresponding window. After normalization, SCG signals from every dataset were resampled to a frequency of 64 Hz, which represents the minimum available sampling frequency among the 3 considered datasets. Resampling is needed to enable a cross-dataset experimental setting, which represents our main contribution. 

In our work, the potential of a DL model was leveraged to detect every systolic complex in a portion of SCG of arbitrary length. This was done by adopting a U-Net \cite{Ronneberger15}, which represents the state-of-the-art solution for semantic segmentation problems. The contracting path of our U-Net contains 4 encoding blocks, each one composed by 2 sequences of \emph{Conv1D, BatchNorm} and \emph{ReLu Activation} layers, followed by a \emph{MaxPooling1D} layer. The expansive path follows a similar architecture where the \emph{MaxPooling1D} layers are replaced by \emph{UpSampling1D} to increase the resolution such that the dimensions of network output equal the input. The U-net is trained for binary semantic segmentation using the annotated systolic complexes.

Our U-Net receives as input an SCG portion of arbitrary length and returns, at each input sample, the probability it belongs to a systolic complex, as shown in Figure \ref{fig:ex}. Having a probability estimate as output, it is possible to obtain a binary mask, where 1 represent samples in a systolic complex, by simply thresholding. Selecting the right threshold represented a fundamental decision in our experiments. In fact, too-low threshold values result in large false positive rates, while a too-high threshold value might produce false negatives. Our U-Net was trained until convergence with an early stopping callback on the validation loss. A batch size of 32 was selected, adopting the binary cross-entropy as loss function and Adam with an initial learning rate of 1e-4 as optimizer.

\begin{figure}
    \centering  \includegraphics[width=0.45\textwidth]{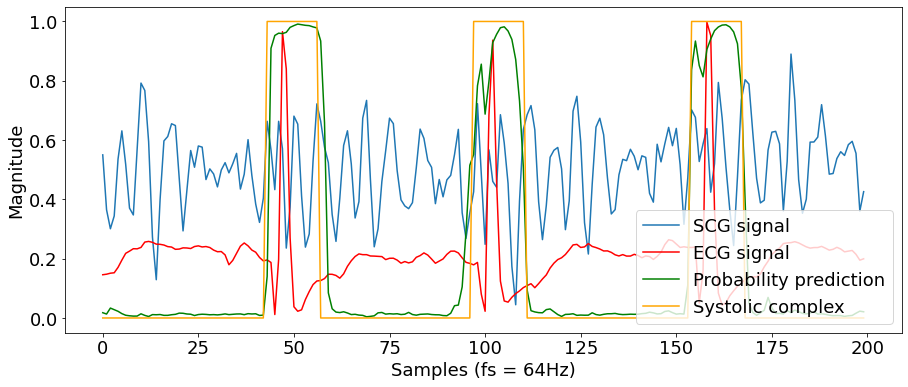}
    \vspace{-2mm}
    \caption{Example of Input (SCG signal), Output (probability prediction) and Ground Truth (Systolic complex) for our DL model. The SCG portion reported here is 
    from BioPoli dataset, thus recorded in uncontrolled conditions.}
    \label{fig:ex}
    \vspace{-0.54cm}
\end{figure}

\section{Results and Discussion}
Our considered datasets present significant differences between each other, as they vary in the sensor equipment, sampling frequency and position of subjects during data collection. These differences poses relevant challenges to DL models which, as far as we know, have never been investigated in the context of SCG analysis. Moreover, we analysed for the first time real-world data collected in uncontrolled conditions, specifically during the sleeping period. For the sake of completeness, our experiments were conducted by testing the effectiveness of both single-channel and multi-channels input signals. All reported results represent the average on the users, which were all considered individually in the testing set.  

We assess the performance of our solution by measuring the number of True Positive (TP), False Positive (FP) and False Negative (FP). In particular, we consider a TP when a bounding box returned by the U-Net contains an AO fiducial point. We then computed the Precision ($\frac{TP}{TP+FP}$), the Recall ($\frac{TP}{TP+FN})$ and the F1-score as below:
\begin{equation}
\notag
F1 = \frac{2*Precision*Recall}{Precision+Recall} = \frac{2*TP}{2*TP+FP+FN}
\end{equation}

\subsection{Single-dataset experiments}
Firstly, we analysed the \emph{single-dataset} experimental setting, where both training and testing data belong to the same dataset. We adopted a Leave-One-Subject-Out (LOSO) cross-validation strategy and we reported averaged results to avoid any subject bias. Our U-Net was trained until convergence, using an early stopping callback on the validation set to restore the weights of the best model, thus avoiding to overfit on training data. The pretrained model was then tested on the subject left out during the training.  
As can be noticed from Table \ref{tab:singleDataset}, the highest performances were reached on CEBS and MEC datasets, which contain healthy subjects data recorded in a controlled environment. 

For the MEC and the BioPoli datasets, we report the results obtained using both a single-channel and a multi-channels approach. In the single-channel experiments we chose the z-axis of the accelerometer, which records the dorso-ventral movements of the chest, which is known to be the most informative channel of the SCG signal. 
As shown in Table \ref{tab:singleDataset}, the lowest performance corresponds to the BioPoli single-channel dataset, but this improves significantly adopting a multi-channels approach. This result confirms that real-world data are the most challenging ones, and that multiple channels are required to cope with that. In contrast, the multi-channel approach is not resulting in a performance boost in the MEC dataset, which contains recordings collected in a controlled scenario.

\subsubsection{Personalization}
The impact of a personalization strategy in the single-dataset setting was analyzed performing a fine-tuning of the pretrained model on user's specific data. Table \ref{tab:persDatasets} shows performances before and after personalization for the 3 considered datasets (both single-channel and multi-channels). As expected,  personalization improves all the cases analyzed in terms of F1-score. We remark that fine-tuning and personalization coincide in the single-dataset setting, as we adopted a LOSO strategy.

\begin{table}[t]
    \centering
    \renewcommand{\arraystretch}{1.2}
    \begin{tabular}{ccccc}
  \multicolumn{2}{c}{}&   \multicolumn{3}{c}{\small{Test statistics}}\\
        & \small{Channels} &\small{Precision} & \small{Recall} & \small{F1-score}

        \\\cline{2-5}
        \multirow{3}{*}{{\rotatebox[origin=c]{90}{MEC}}} 
        &   \small{Single (z-axis Acc.)} & 0.93 &0.92 & 0.92 \\        
        &   \small{3 Acc.} & 0.92 &0.92 & 0.92 \\ 
        &   \small{3 Gyr.} & 0.92 & \textbf{0.93} & 0.92 \\ 
        &   \small{3 Acc. + 3 Gyr} & \textbf{0.93} & 0.92 & \textbf{0.92} \\ 
        \cline{2-5} 
        \multirow{3}{*}{{\rotatebox[origin=c]{90}{BioPoli}}} 
        &   \small{Single (z-axis Acc.)} & 0.90 &0.87 & 0.88 \\        
        &   \small{3 Acc.} & 0.92 &0.90 & 0.91 \\ 
        &   \small{3 Gyr.} & 0.93 & 0.90 & 0.91 \\ 
        &   \small{3 Acc. + 3 Gyr} & \textbf{0.93} & \textbf{0.92} & \textbf{0.92} \\ 
        \cline{2-5}
        \multirow{1}{*}{{\rotatebox[origin=c]{90}{CEBS}}} 
        &   \small{Single (z-axis Acc.)} & 0.95 & 0.93 & 0.94 \\        
    \end{tabular}
    \vspace{3mm}
    \caption{Precision, Recall and F1-score in the single-dataset setting. For MEC and BioPoli datasets are reported the scores of both single-channel and multi-channels approaches.} 
    \label{tab:singleDataset}
\end{table}

\begin{table}[t]
   \centering
    \begin{tabular}{cccc} 
    & \multicolumn{2}{c}{F1-score} \\ 
        Dataset & Initial  & Personalized & $\Delta \%$ \\ \hline \hline
        CEBS &  \textbf{0.94}       &  \textbf{0.95} & + 1\% \\ \hline
        MEC (single)   &0.92         & 0.92 & -  \\
        MEC (multi)   & 0.92         & 0.93  & +1\%  \\ \hline
        BioPoli (single)   &0.88         &  0.93 & \textbf{+5\%} \\
        BioPoli (multi)  &0.92         & \textbf{0.95} & +3\% \\ \\
    \end{tabular}
    \vspace{-1mm}
    \caption{Comparison between the F1-score of the Initial and Personalized models over the 3 datasets.}
    \label{tab:persDatasets}
    \vspace{-7mm}
\end{table}

\vspace{-1mm}
\subsection{Cross-dataset experiments}
\vspace{-0.8mm}
A more challenging scenario was investigated, where train and test users belong to different datasets. In particular, the same model pretrained on the full training dataset was tested individually on each user of the second dataset.  
The results averaged among the users are reported in Table \ref{tab:acc_cross}. We notice that the best cross-dataset result was obtained for the MEC-CEBS case. This result highlights that our DL solution is effective even in a cross-dataset scenario when source and target data are both collected in a controlled environment, i.e. when the user's movements are limited during the recording protocols. The MEC-CEBS experiment even achieves superior performance than the CEBS-CEBS one: this confirms that there is no significant domain shift between controlled data from the two public datasets and that the differences are possibly related to other factors, such as the cardinality of the training set. The lowest performances are achieved when testing on the BioPoli dataset, as shown in Table \ref{tab:acc_cross}. This proves that training over clean SCG signals is not so effective for testing in a real-world scenario. On the contrary, a DL model trained on real-world data can extract the relevant features to correctly identify a systolic complex in simpler scenarios. This is confirmed by the results in Table \ref{tab:acc_cross}, when the BioPoli data are selected as training set. Additionally, Table \ref{tab:acc_cross} shows that training on multi-source datasets always improves the performance of our model, possibly as this promotes the network to learn domain invariant features.

\begin{table}[t]
   \centering
    \begin{tabular}{cc|ccc} 
        Train &Test & Precision  & Recall  & F1-score \\ \hline \hline
        MEC &MEC (single) &0.93 &0.92 &0.92 \\
        MEC &MEC (multi) &0.93 &0.92 &0.92 \\
        MEC &CEBS  &  \textbf{0.95}       & \textbf{0.97}  & \textbf{0.95}  \\
        MEC &BioPoli (single)  &0.86         &  0.82  &  0.84 \\ 
        MEC &BioPoli (multi)  & 0.83        & 0.79   & 0.81  \\ \hline
        CEBS &CEBS  & \textbf{0.95}          & \textbf{0.93}   &  \textbf{0.94} \\
        CEBS &MEC  & 0.90          & 0.89   &  0.89 \\
        CEBS &BioPoli & 0.84          & 0.82    &  0.83 \\ \hline
        BioPoli &BioPoli (single) & 0.90          & 0.87    &  0.88 \\
        BioPoli &BioPoli (multi) & 0.93          & 0.92    &  0.92 \\
        BioPoli &MEC (single) & 0.91          & 0.87    &  0.88 \\
        BioPoli &MEC (multi) & \textbf{0.94}          &  \textbf{0.93}   & \textbf{0.93}  \\
        BioPoli &CEBS & 0.88          & 0.88    &  0.88 \\ \hline
        MEC+CEBS &BioPoli & 0.87         & 0.86    & 0.86  \\ 
        MEC+BioPoli &CEBS & \textbf{0.95}         & \textbf{0.98}    & \textbf{0.96}  \\ 
        BioPoli+CEBS &MEC & 0.91         & 0.91    & 0.91  \\ \\

    \end{tabular}
    \vspace{-1mm}
    \caption{Precision, Recall and F1-score achieved in the cross-dataset setting. For the combinations not involving CEBS dataset we report both the single-channel and multi-channels results.}    
    \label{tab:acc_cross}
\end{table}

\begin{table}[t]
   \centering
    \begin{tabular}{cccc} 
    & \multicolumn{3}{c}{F1-score} \\ 
        Train-Test & Initial  & Fine-Tuned  & Personalized \\ \hline
        BioPoli-BioPoli &  \textbf{0.88}       & -  & \textbf{0.91}  \\
        CEBS-BioPoli   &0.83         &  0.89  &  0.89 \\
        MEC-BioPoli  &0.84         &  0.88  &  0.89 \\ 
        CEBS+MEC-BioPoli  &0.86         &  \textbf{0.90}  &  \textbf{0.91} \\ \\

    \end{tabular}
    \vspace{-1mm}
    \caption{F1-score of Initial, Fine-Tuned and Personalized model adopting BioPoli as testing dataset (single-channel).}
    \label{tab:persBio}
    \vspace{-9mm}
\end{table}

\subsubsection{Fine-tuning and Model Personalization}
Finally, we compare the effectiveness of traditional fine-tuning and model personalization to counteract domain shift in a \emph{cross-dataset} experiment. The BioPoli dataset was chosen as testing set, since this represented the most challenging scenario. In the traditional fine-tuning, the testing dataset was split again in two subsets, the first used for the fine-tuning of the model and the second for the performance assessment. In personalization, the initial model was fine-tuned on a portion of data from a single user and then tested on remaining data of the same user. Table \ref{tab:persBio} shows that both fine-tuning and personalization improved the performance of the initial model trained on the training set, with the latter achieving slightly superior performance. In particular, the best result was obtained by personalizing the model pretrained on the BioPoli dataset itself. This confirms the domain shift between data recorded in an ideal scenario with respect to real-world signals.

\section{Conclusions}
In this work we faced the task of ECG-free SCG systolic complex identification adopting a state-of-the-art DL solution. We performed an extensive experimental analysis to assess the performance of our model in a cross-dataset scenario, considering also real-world data. Our experiments demonstrated the limitations of DL models trained exclusively on SCG signals collected in a controlled environment. Real-world SCG signals turns out to be more challenging to analyze than those in publicly available datasets, and  require model personalization or fine tuning strategies. Finally, we also showed the benefits of learning over multiple sensors' channels, which improved the performance of our model in both single-dataset and cross-dataset settings.
\vspace{-1.2mm}
\bibliographystyle{IEEEtran}
\bibliography{embc}

\end{document}